\documentclass[a4paper,fleqn]{cas-dc}
\usepackage[numbers,sort&compress]{natbib} 
\usepackage{graphicx} 
\usepackage{float}
\usepackage{textcomp,mathcomp}
\usepackage{subcaption}
\usepackage{amsmath}
\usepackage{soul}
\usepackage{color, xcolor} 
\usepackage{rotating} 
\usepackage{diagbox} 
\usepackage{ulem}
\usepackage{tabularx} 
\usepackage{diagbox}  
\def\tsc#1{\csdef{#1}{\textsc{\lowercase{#1}}\xspace}}
\tsc{WGM}
\tsc{QE}
\tsc{EP}
\tsc{PMS}
\tsc{BEC}
\tsc{DE}
\begin{document}
\let\WriteBookmarks\relax
\def\floatpagepagefraction{1}
\def\textpagefraction{.001}
\shorttitle{Junction Temperature and Position Prediction in $3$D HBM Chiplets}
\shortauthors{C ZHANG et~al.}

\title [mode = title]{Neural Network Surrogate Model for Junction Temperature and Hotspot Position in $3$D Multi-Layer High Bandwidth Memory (HBM) Chiplets under Varying Thermal Conditions} 

\tnotetext[1]{The project is supported by the Peng Cheng Laboratory, Peng Cheng Cloud-Brain, and the Center for Computational Science and Engineering at Southern University of Science and Technology.}


\author[1,2]{Chengxin ZHANG}[]

\address[1]{Pengcheng laboratory, Shenzhen, China}

\author[1]{Yujie LIU}[%
 ]
\cormark[1]
\ead{liuyj02@pcl.ac.cn}


\address[2]{School of Microelectronics, Southern University of Science and Technology, Shenzhen, China}

\author[2]{Quan CHEN}
\cormark[1]
\ead{chenq3@sustech.edu.cn}

\cortext[cor1]{Corresponding author}


\begin{abstract}
As the demand for computational power increases, high-bandwidth memory (HBM) has become a critical technology for next-generation computing systems. However, the widespread adoption of HBM presents significant thermal management challenges, particularly in multilayer through-silicon-via (TSV) stacked structures under varying thermal conditions, where accurate prediction of junction temperature and hotspot position is essential during the early design. This work develops a data-driven neural network model for the fast prediction of junction temperature and hotspot position in 3D HBM chiplets. The model, trained with a data set of $13,494$ different combinations of thermal condition parameters, sampled from a vast parameter space characterized by high-dimensional combination (up to $3^{27}$), can accurately and quickly infer the junction temperature and hotspot position for any thermal conditions in the parameter space. Moreover, it shows good generalizability for other thermal conditions not considered in the parameter space. The data set is constructed using accurate finite element solvers. This method not only minimizes the reliance on costly experimental tests and extensive computational resources for finite element analysis but also accelerates the design and optimization of complex HBM systems, making it a valuable tool for improving thermal management and performance in high-performance computing applications.
\end{abstract}


\maketitle

\section{Introduction}

As information technology continues to advance at an unprecedented pace, the need to improve the performance and integration density of integrated circuits (IC) has become a central driver of industrial progress \cite{claasen2006industry}. To meet the growing demands for computational power and memory bandwidth, high-bandwidth memory (HBM) technology has emerged as a critical solution \cite{kim2022overview}. HBM enables faster data transfer between memory and processing units, making it indispensable for high-performance computing (HPC), AI accelerator, and data center applications \cite{mukhopadhyay2019heterogeneous}. However, the integration of HBM in $3$D stacked configurations introduces significant thermal management challenges, particularly within multilayer Through-Silicon-Via (TSV) structures \cite{lau2014overview}. The thermal performance of TSV is vital not only for the stability of HBM modules but also for the overall performance and reliability of the entire chip stack \cite{gambino2015overview}. In this context, efficient thermal management is essential to ensure that HBM technology can operate within safe thermal limits, thereby maintaining high performance in demanding computing environments \cite{tang2024brief}. The thermal management challenges associated with HBM systems are multifaceted, arising from factors such as high power density, limited thermal conductivity in vertical stacks, and the complex interaction between stacked layers and TSV \cite{kim2023thermal}. The high power density within stacked chips creates localized hot spots that can cause thermal stress and performance degradation if not properly managed. Furthermore, the thermal conductivity of the materials used in TSV is often insufficient, resulting in a build-up of heat that adversely affects the entire system \cite{farmahini2018challenges, mccann2018warpage}. These challenges require effective thermal design solutions that require the integration of advanced predictive techniques to ensure optimal thermal performance. 

Traditional Finite Element Analysis (FEA) is widely used for its ability to simulate heat conduction processes with high precision\cite{wang2024heat}, but suffers from significant computational overhead when applied to large-scale complex systems such as $3$D stacked HBM \cite{kim2023thermal}. This computational burden makes it challenging to perform rapid design iteration and optimization. Faster algorithms are often used in such cases, but these tend to rely on simplified models that may compromise accuracy in complex $3$D structures. As a result, the ability to optimize thermal performance in advanced HBM architectures is significantly hindered, especially when designing high-density chip stacking \cite{mccann2018warpage} and complex TSV configurations \cite{farmahini2018challenges}. The complexity of these thermal dynamics necessitates novel approaches to enhance predictive accuracy and optimize thermal behavior \cite{shen2023thermal, khadirsharbiyani2024comprehensive}. AI-driven models, particularly neural networks, have shown great potential in predicting thermal performance \cite{ozsunar2009prediction} and hotspot positions \cite{wang2024heat} in complex systems. Data-driven approaches, particularly neural network (NN) models, have gained significant attention due to their ability to handle large datasets and complex nonlinear relationships efficiently \cite{10247998, wang2022chip, wang2024heat, narayana2016artificial}. These models can predict thermal behavior with high accuracy by learning from historical simulations or experimental data, providing a faster alternative to traditional methods. By training on FEA-generated data, neural networks can offer predictions of thermal performance and hotspot positions with minimal computational overhead compared to full-scale simulations.

\begin{figure*}
	\centering
	\includegraphics[width=0.9\textwidth]{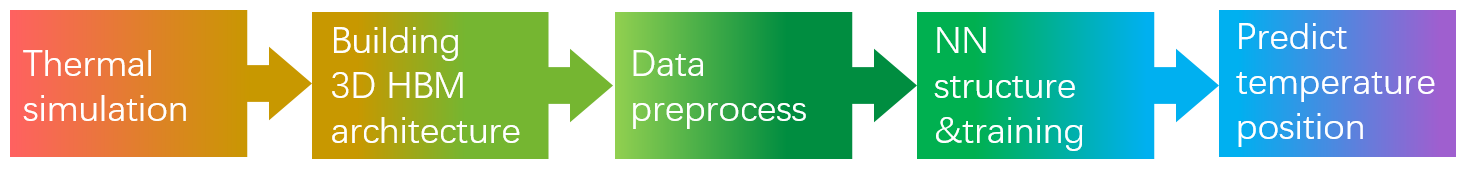}
	\caption{Flowchart of the process for 3D HBM chiplets junction temperature and position prediction neural network model construction.}\label{pic:3}
\end{figure*}

\begin{figure}
	\centering
	\includegraphics[width=0.4\textwidth]{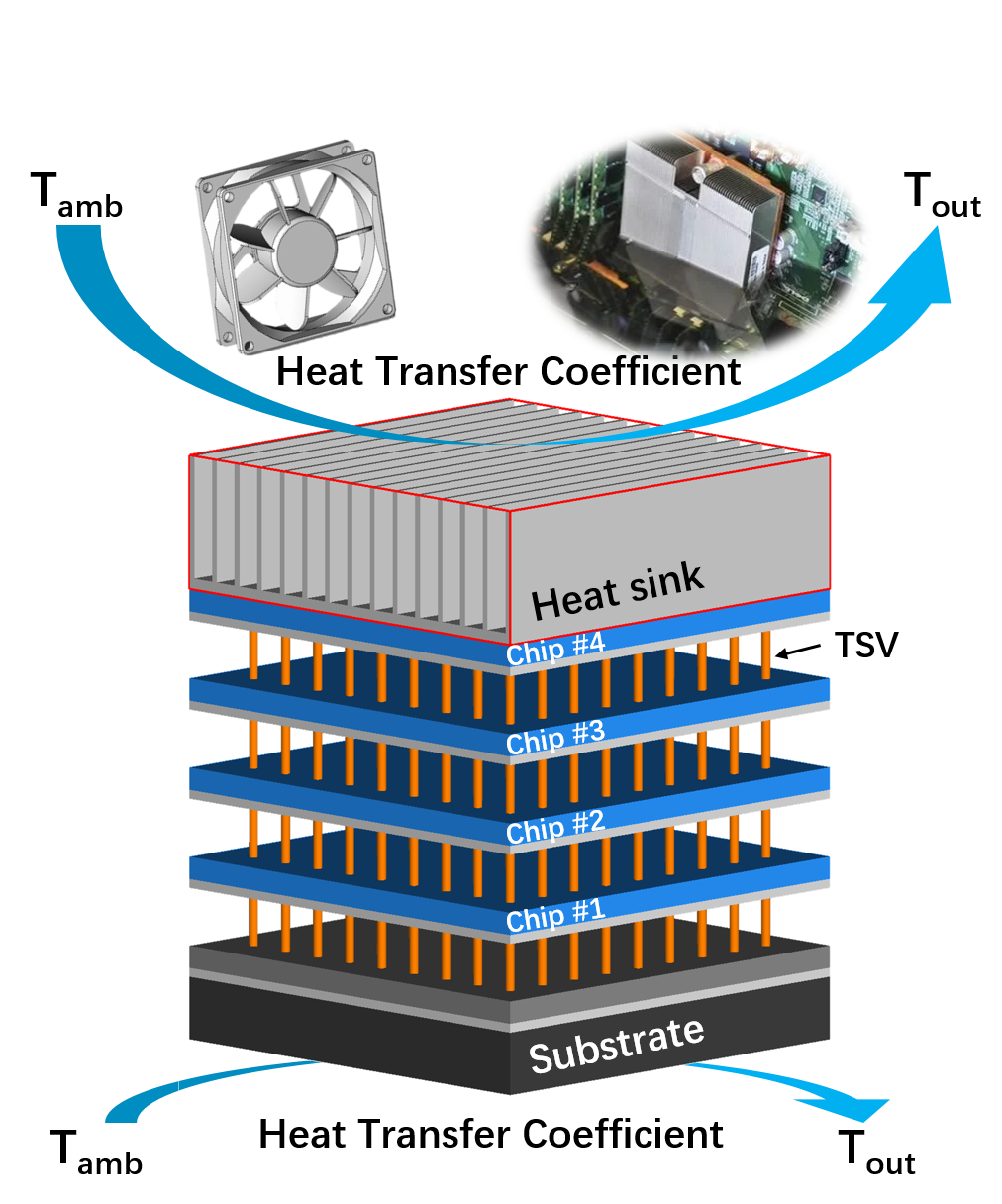}
	\caption{Schematic diagram of a 4HBM chiplet structure.}\label{pic:q}
\end{figure}

This study focuses on developing a surrogate neural network thermal prediction model for $3$D HBM structures under varying conditions, with a particular emphasis on real-time prediction and mitigating repetitive calculations often encountered during early design. As shown in fig.\ref{pic:3}, the process begins with numerical modeling of the HBM structures, followed by dosens of thermal simulations to obtain junction temperature and hotspot position. The simulation results are then preprocessed and fed into a designed neural network for training, which ultimately produces the surrogate model. The structure of this paper is organized as follows. Section $2$ introduces the HBM structures and thermal simulations with different parameters used to generate data for neural network training. Section $3$ discusses the design of the NN model, including data pre-processing, NN structure, and the training process. Section $4$ analyzes the experimental results and Section 5 concludes.

\section{HBM architecture and thermal simulation}

This section aims to support NN training by providing substantial training data through numerical simulations. Although the research encompasses various HBM structures (including $1$-layer, $2$-layer, $4$-layer, and $8$-layer), only $4$-layer HBM ($4$HBM) and $8$-layer HBM ($8$HBM) configurations will be primarily discussed for clarity and representational purposes. This section is divided into three parts: $2.1$ $3$D HBM architecture; $2.2$ thermal simulation; $2.3$ parameter combination space.

\subsection{$3$D HBM architecture}
 
 \begin{figure}
 	\centering
 	\includegraphics[width=0.48\textwidth]{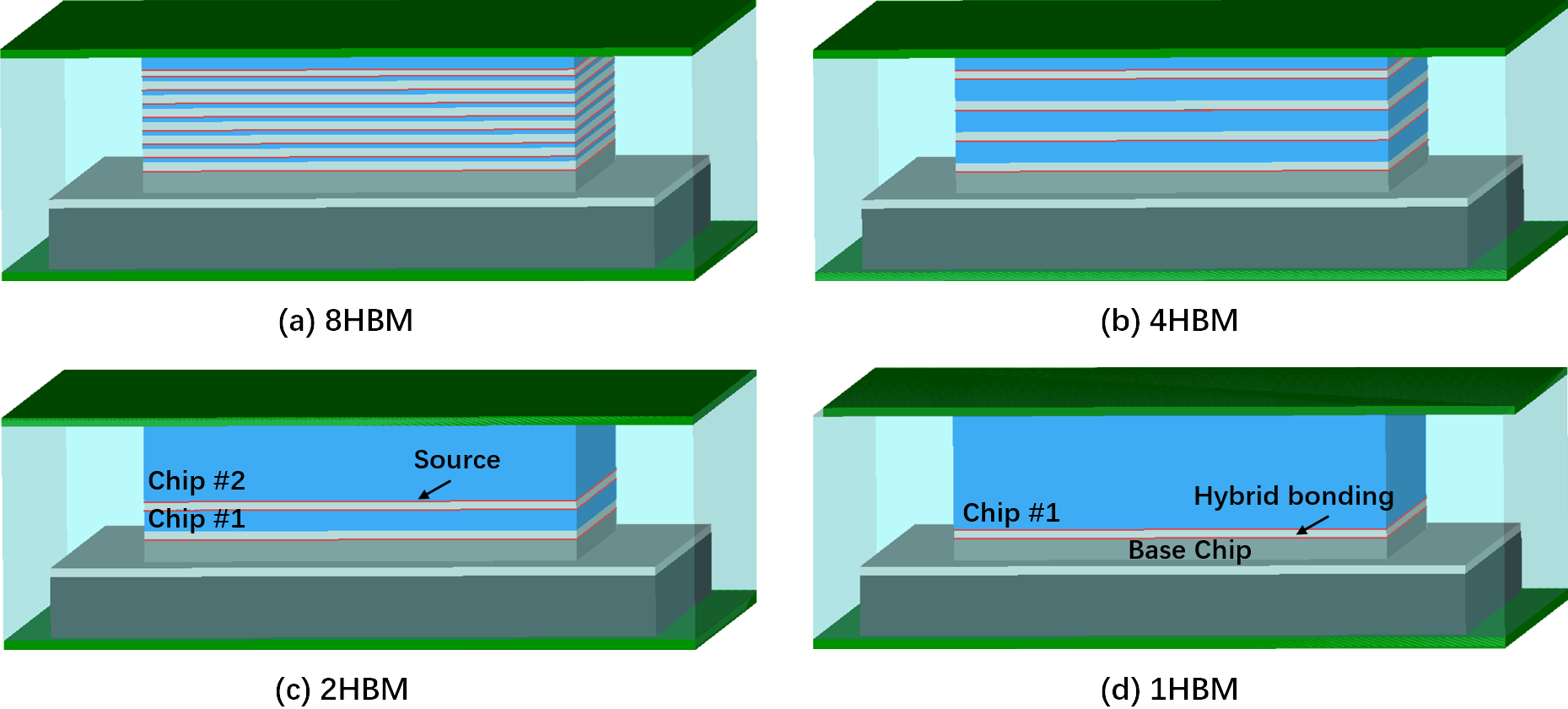}
 	\caption{Schematic diagram of boundary condition application regions for mult-layer HBM chiplets.}
 	\label{pic:2}
 \end{figure}
 
As illustrated in fig. \ref{pic:q}, this section employs a $4$HBM structure as an exemplar for structural explanation. From bottom to top, the $4$HBM structure is composed of a silicon substrate in the bototm, followed by $4$ HBM layers sequentially stacked, each containing internal TSV structures interconnected with adjacent layers through bonding layers, and at the top surface it is equipped with a cooling condition. Cooling condition including air cooling, liquid cooling, and phase-change cooling, implemented through different heat transfer coefficient (HTC) parameters. 

As shown in fig. \ref{pic:2}, this schematic systematically demonstrates $1$-layer, $2$-layer, $4$-layer, and $8$-layer HBM stacking architectures. Red plane explicitly mark the spatial distribution of heat sources (power), while the blue blocks represent multi-layer chip structures. Notably, all chip layers except the top tier contain high-density TSV arrays, which serve as critical thermal conduction paths while enhancing interconnect bandwidth. The green plane represents HTC boundary conditions that exhibit temperature-dependent nonlinearity.

\subsection{Thermal simulation} 

\begin{figure*}
	\centering
	\includegraphics[width=1\textwidth]{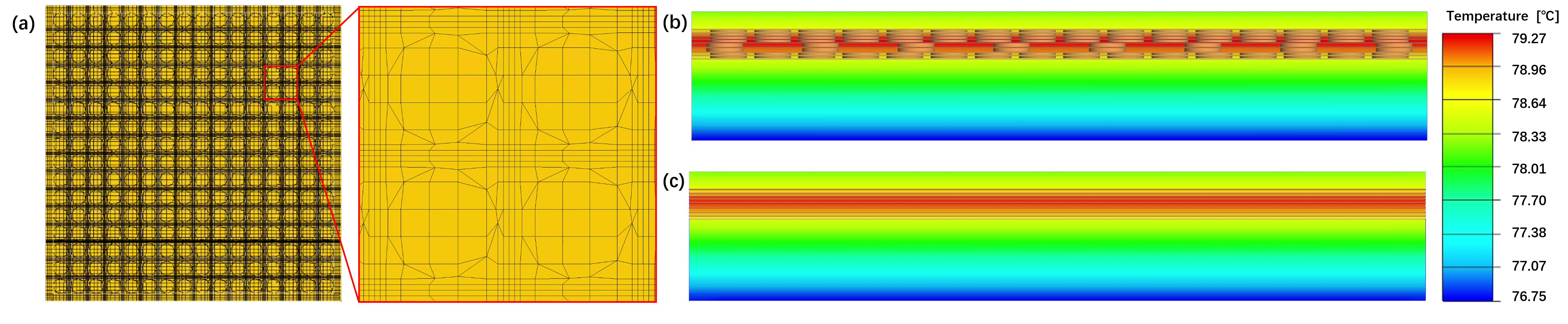}
	\caption{(a): Horizontal (x-y) plane mesh and localized mesh details of the 8HBM before equivalence; (b) and (c): Temperature distributions in the x-z plane before and after TSV equivalence.}\label{pic:4}
\end{figure*}

The thermal simulations are conducted using the ANSYS software, by using python scripts to automatically process the simulation for different combination of HBM structure and thermal parameters. Taking an 8HBM structure as an example, thermal simulations were performed on the models before and after TSV processing, and the results were compared (see fig.~\ref{pic:4}). As shown in figure (a), the x-y-plane mesh view on the left illustrates the dense TSV layer network. These meshes are non-uniformly distributed in the horizontal direction, which increases computational complexity. To reduce modeling time and computational resources and accelerate mesh generation, the thermal conductivity of TSV within the chip is equivalently processed \cite{wang2024121499}. This method simplifies the complex structure of TSV, effectively averaging the thermal properties of the TSV and surrounding materials. The following equations describe the thermal conductivity in the x-y plane \(k_{xy}\) and the z direction \(k_{z}\) for the multilayer HBM structure: Let the area of the die be $S_{\text{die}} = A \times B$, with individual TSV radius $r$ and total count $N$. The TSV areal density is defined as

\begin{flalign}
	\rho = \frac{N \pi r^2}{AB}& \label{eqn-x} 
\end{flalign}
The equivalent square region edge length occupied by each TSV in the plane is derived as:
\begin{flalign}
	c_x = c_y = \sqrt{\frac{AB}{N}} = \sqrt{\pi r^2 / \rho}& \label{eqn-w} 
\end{flalign}
Based on series-parallel thermal resistance theory, the in-plane equivalent thermal conductivity ($k_{xy}$) is formulated as:
\begin{flalign}
	k_{xy} = \frac{B \cdot k_{\text{Si}} \cdot \left[ c_x k_{\text{Cu}} + \frac{(A - c_x)k_{\text{Si}}}{A} \right]}{k_{\text{Si}} c_y + (B - c_y) \left[ c_x k_{\text{Cu}} + \frac{(A - c_x)k_{\text{Si}}}{A} \right]} & \label{eqn-y} 
\end{flalign}
The vertical heat flux path comprises parallel contributions from TSV arrays and silicon substrate:
\begin{flalign}	
	k_z = \frac{\pi r^2 N k_{\text{TSV}} + \left( S_{\text{die}} - \pi r^2 N \right) k_{\text{Si}}}{S_{\text{die}}} & \label{eqn-z}
\end{flalign}
where \( A \) and \( B \) are the side lengths of chip. Equations \eqref{eqn-x} - \eqref{eqn-w} calculate the effective radius \( c_x \) and \( c_y \) for a circular area, where \( N \) is the number of TSV and \( r \) is the radius. Equation \eqref{eqn-y} calculates the horizontal thermal conductivity \( k_{xy} \), where \( k_{\text{Si}} \) is the thermal conductivity of silicon, \( k_{\text{Cu}} \) is the thermal conductivity of copper, and \( A \) and \( B \) are the side lengths of the chip. Equation \eqref{eqn-z} calculates the vertical thermal conductivity \( k_z \), where \( k_{\text{TSV}} \) is the thermal conductivity of TSV, \( S_{\text{die}} \) is the chip area and \( S_{\text{TSV}} \) is the TSV area. 

\begin{figure}
	\centering
	\includegraphics[width=0.45\textwidth]{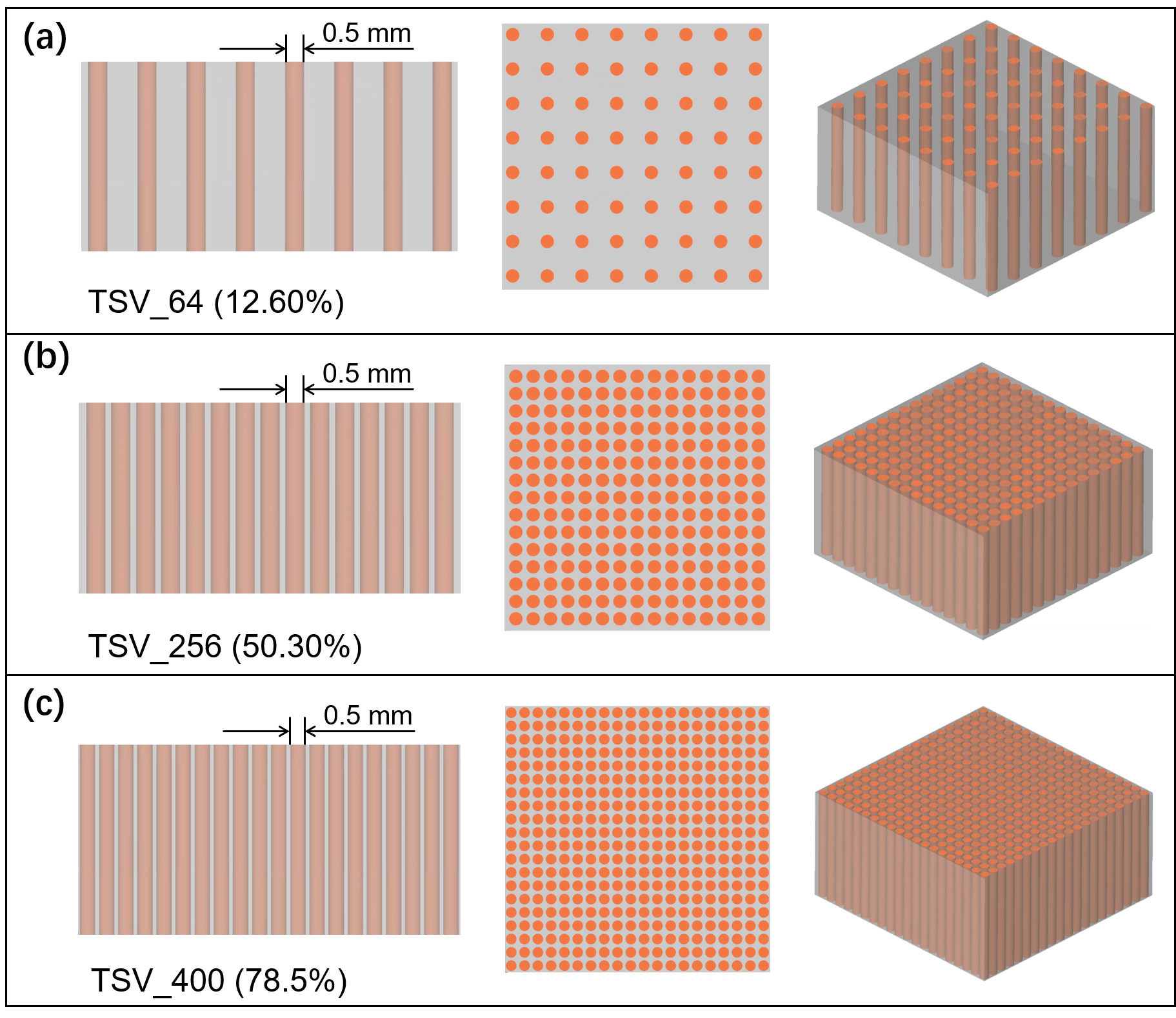}
	\caption{Before TSV equivalence model in simulations.}\label{pic:1}
\end{figure}

\begin{table}
	\centering
	\caption{Temperature results for before and after TSV equivalent.}\label{tab:t2}
	\setlength{\tabcolsep}{9mm} 
	\setlength{\tabcolsep}{1.2mm} 
	\begin{tabular}{llll}
		\hline
		~ 								& Tmax 		& Tmin 			& position \\ \hline
		TSV ($\tccentigrade $)							& 79.27		& 76.75 		& Chip4 \\ 
		TSV equivalent ($\tccentigrade $) 					& 79.17 	& 76.77 		& Chip4 \\
		TSV-TSV equivalent ($\tccentigrade $) 				& 0.10 		& -0.019		& -- \\ 
		(TSV-TSV equivalent)/TSV 		& 0.13$\%$	& -0.025 $\%$ 	& -- \\ \hline
	\end{tabular}\\
\end{table}

Since the equivalence process eliminates the need for detailed TSV modeling, the number of simulation objects before equivalence is significantly larger than after equivalence, thereby substantially reducing the computation time. The number of meshes before equivalence is $1,779,790$, while after equivalence it is $32,634$. From the temperature distribution before equivalence (figure (b)) and after equivalence (figure (c)), it can be observed that the temperature distribution regions are consistent. Analysis of the results shows that the discrepancies between the two sets of simulation outcomes are negligible, as supported by the specific data provided in tab. \ref{tab:t2}. This indicates that the TSV equivalent model has a minimal impact on the overall temperature distribution. Consequently, these results validate the simplification of models for this application, as the accuracy remains largely unaffected. The junction temperature is $79.27 \tccentigrade $ before equivalence and $79.17\tccentigrade $ after equivalence, with the hot spot located at chip4 in both cases, suggesting that the equivalence process has minimal impact on the junction temperature and the hotspot location.

\begin{figure}
	\centering
	\begin{subfigure}[b]{0.45\textwidth}
		\centering
		\includegraphics[width=\textwidth]{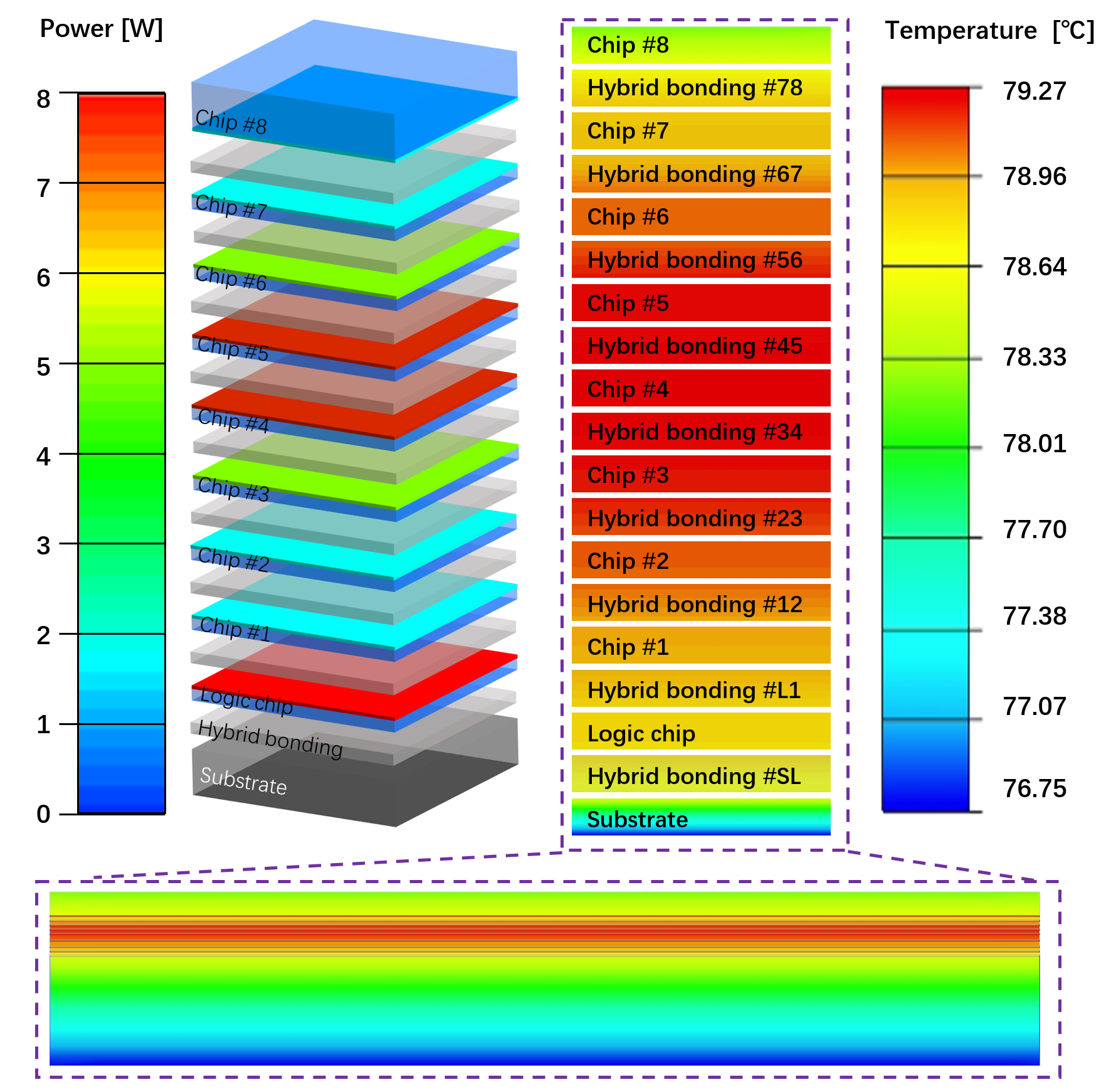}
		\label{fig:temp-dist}
	\end{subfigure}
	\begin{subfigure}[b]{0.45\textwidth}
		\centering
		\includegraphics[width=\textwidth]{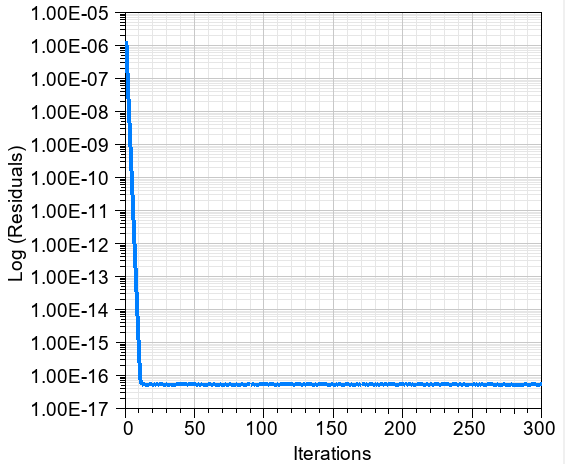}
		\label{fig:residual-trend}
	\end{subfigure}
	\caption{8HBM thermal simulation results: (top) temperature distribution of 8HBM chiplets and (bottom) residual trend during the iterative process.}
	\label{pic:D}
\end{figure}

Fig. \ref{pic:D} (top) presents the thermal simulation result incorporating parameters that closely reflect a practical application scenario, including the chip thicknesses are $30$ µm for layer $1$, $50$ µm for layers $2$-$3$, $30$ µm for layers $4$-$7$, and the remaining thickness for layer $8$ being $235$ µm to achieve a total thickness of $720$ µm, with the top chip thickness. Thermal conductivity is identified by the number of copper pillars in each chip, e.g., chip$1$ uses $256$ pillars, resulting in pillar counts for chip$1$-$8$ of $256$, $256$, $64$, $64$, $256$, $256$, $256$, $256$, $0$, where $0$ denotes pure silicon. The power values for chip$1$-$8$ are $2$ W, $2$ W, $5$ W, $8$ W, $8$ W, $5$ W, $2$ W, $2$ W, respectively. A HTC of $4,000$ W/m²·K is selected for both the top surface and the substrate. Fig. \ref{pic:D} (bottom) illustrates the residual trend observed during the iterative process of the thermal simulation. The graph shows a rapid decrease in residuals during the initial iterations, followed by steady stabilization, indicating the convergence of the numerical method.

\subsection{Parameter combination space}

\begin{table}
	\centering
	\caption{Simulation scheme parameters.}\label{tab:t1}
	\setlength{\tabcolsep}{1mm} 
	\begin{tabular}{ll}
		\hline
		Physical parameters 				& Data \\ \hline
		Number of layers 					& 1, 2, 4, 8 \\ 
		& \\
		Chip thickness (\textmu $m$) 		& 10, 30, 50 \\ 
		& \\
		Chip (TSV)  						& TSV\_64	(12.6$\%$)://170.85 $\perp$ 179.79\\ 
		thermal conductivity 				& TSV\_256 	(50.3$\%$)://327.37 $\perp$ 242.01 \\ 
		($W/m$·$K$) 						& TSV\_400 	(78.5$\%$)://372.22 $\perp$ 317.49 \\
		& \\
		Substrate  							& //130.00 $\perp$130.00\\
		thermal conductivity & \\
		& \\
		Hybrid bonding   					& //0.70 $\perp$15.19\\ 
		thermal conductivity 				& \\
		& \\
		Power (W)  							& 1HBM: 2 - 8;\\
		~ 									& 2HBM: 2, 5, 8;\\
		~ 									& 4HBM: 2, 5, 8;\\ 
		~ 									& 8HBM: 2, 5, 8;\\ 
		& \\
		Top/Bottom HTC 			& 200 4000 10000 \\ 
		($W/m_2K$) 				& ~\\ \hline
	\end{tabular}\\
	\footnotesize
	\raggedright Forced convection in air: 200 $W/m_2K$;\\
	Forced convection in liquids: 4,000 $W/m_2K$; \\
	Boiling/phase change: 10,000 $W/m_2K$.\\ \par
\end{table}

To ensure practical relevance of the simulations, key parameters and their respective ranges were systematically evaluated (see tab.~\ref{tab:t1} for details). The number of chip layers, ranging from $1$ to $8$, represents the memory chip layers within the HBM stack, significantly influencing thermal distribution and inter-layer thermal coupling. The thickness of each memory chip layer varies from $10$ µm to $50$ µm, with the total stack thickness fixed at $720$ µm, directly affecting temperature rise rates and thermal gradients. Material properties, particularly thermal conductivity, play a pivotal role in determining the heat dissipation efficiency. A schematic representation of the equivalent model is provided in fig.~\ref{pic:1}. For example, in fig.~\ref{pic:1} (b), TSV\_$256$ denotes the arrangement of $256$ TSV copper pillars within the silicon layer, with a volume filling fraction of $50.30$\%, indicating the proportion of copper pillars relative to the silicon area. Power dissipation per chip layer ranges from $2$ W to $8$ W \cite{JESD238A}, covering both peak power during high-speed operations and idle power during standby modes \cite{bae2024thermo}. Environmental factors, including an ambient temperature of $25 \tccentigrade $and a heat transfer coefficient (HTC) ranging from 200 to 10,000 W/m²·K, were incorporated to represent external cooling conditions, which critically influence heat dissipation and overall thermal stability; however, due to the exponential growth in computational requirements and experimental complexity with the increasing number of variables and their ranges, the exhaustive combinatorial analysis of all possible parameter combinations within this vast multidimensional design space is infeasible within a practical time frame.

\section{Junction temprature and hotspot position prediction neural network model}
This section is divided into three parts: data pre-processing, neural network architecture, and neural network training.       

\subsection{Data pre-processing}

\begin{figure*}
	\centering
	\includegraphics[width=1\textwidth]{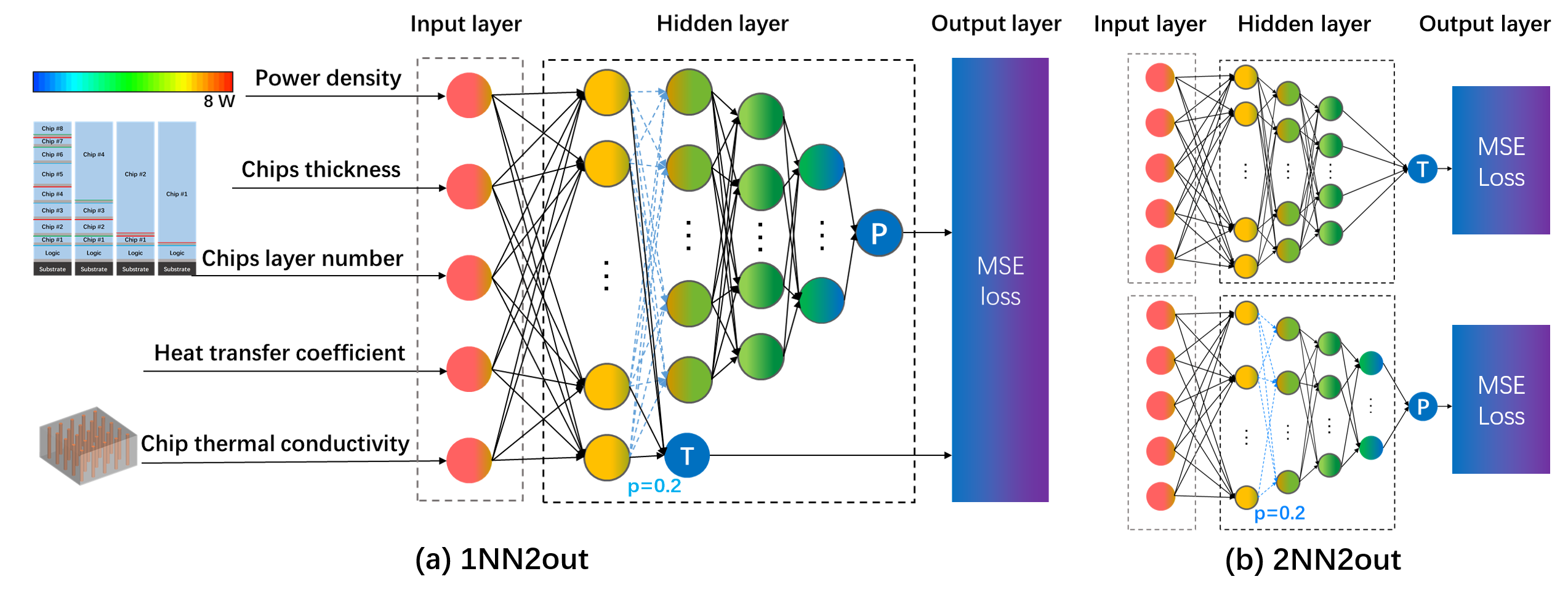}
	\caption{Schematic of the neural network architecture: (a) 1NN2out; (b) 2NN2out.}
	\label{pic:b}
\end{figure*}

The number of parameter combinations involved in exhaustive enumeration grows exponentially with the number of HBM layers, which is computationally infeasible in practice. Using $8$HBM as an example: ($1$) chip thickness of logic layer and the $7$ layer chips (with the top layer filled to $720$ µm) consists of 8 layers; ($2$) Chip (TSV) thermal conductivity of logic layer and the $7$ layer chips (with the top layer being silicon) consists of 8 layers; ($3$) power of logic layer and the $8$ layer chips consists of $9$ layers, each with $3$ possible values. ($4$) HTC for the top and bottom surfaces consist of $2$ sides; The total number of combinations for these variables is $8$ + $8$ + $9$ + $2$ = $27$, resulting in $3^{27}$ possible combinations. To address this issue, partial Latin hypercube sampling (LHS) was employed to sample the parameter space, resulting in $13,494$ valid combinations after excluding nonphysical temperature data, which is significantly reduced compared to exhaustive enumeration, thereby drastically shortening the simulation time. Tab. \ref{tab:t4} compares the number of combinations of parameters and the usage of data between the exhaustive enumeration simulation and the partial LHS simulation.

To prepare the simulation datasets for NN training, the data undergoes pre-processing before being input into the network. First, normalization is applied to ensure consistency across input variables, with scaling factors as detailed in tab. \ref{tab:scaling_factors}. Placeholder values of $-1$ are used to align data dimensions where necessary. Subsequently, the datasets is randomized to enhance diversity during training and is split into an $8 : 2$ ratio for training and testing sets, respectively.

\begin{table}
	\centering
	\caption{Comparison of simulation schemes and data usage.}\label{tab:t4}
	\setlength{\tabcolsep}{1.2mm} 
	\begin{tabular}{lll}
		\hline
		Parameter combinations 									& number \\ \hline
		Exhaustive enumeration simulation 		 				& 	$7,625,611,854,306$  \\ 
		\quad \quad 1HBM 						 				& 	$3^{4}*7 = 567$  \\ 
		\quad \quad 2HBM 						 				& 	 $3^{9} = 19,683$ \\ 
		\quad \quad 4HBM 					 					& 	  $3^{15} = 14,348,907$ \\ 
		\quad \quad 8HBM 					 					&  $3^{27} = 7,625,597,484,987$\\ \hline
		Partial LHS simulation 	& 	  \\ 
		\quad \quad Simulation dataset   			 			& 13,494	  \\ 
		\quad \quad (Excluding non-physical T)					& 	  \\ 
		\quad \quad Full datasets 			 					& 	13494   \\ 
		\quad \quad Partial datasets 			 				& 	2000   \\\hline 
	\end{tabular}\\
\end{table}

\begin{table}[htbp]
	\centering
	\caption{Scaling factors for normalization.}
	\setlength{\tabcolsep}{8mm} 
	\label{tab:scaling_factors}
	\begin{tabular}{ll}
		\hline
		Parameter        		& Scaling factor \\ \hline
		Number of layers       & 100                \\ 
		Chip thickness         & 4000               \\ 
		Power                  & 100                \\ 
		HTC                    & 0.3                \\ 
		Junction temperature   & 2                  \\ 
		Hotspot position       & 1000               \\ \hline
	\end{tabular}
\end{table}

\subsection{Neural network architecture}

As shown in fig. \ref{pic:b}, two different neural network architectures are considered for the prediction of junction temperature and hotspot position. The first architecture, named $1$NN$2$out, applies a single neural network with two output structures to enhance computational efficiency and reduce redundant parameters. It comprises four hidden layers, with the junction temperature prediction output after the first hidden layer. The hotspot position prediction branch incorporates a dropout layer (p=$0.2$) after the first hidden layer and generates its output after the fourth hidden layer. The detailed architecture is illustrated in fig. \ref{pic:b}(a), with hidden layer configurations of $512$, $256$, $128$, and $64$ neurons, respectively.
The second architecture, named $2$NN$2$out (fig. \ref{pic:b}(b)), employs two separate neural networks: one for junction temperature prediction with three hidden layers ($512$, $256$, $128$ neurons) and linear output activation, and another for hotspot position prediction with four hidden layers ($512$, $256$, $128$, $64$ neurons), using nonlinear output activation and a dropout layer (p=$0.2$) after the first hidden layer.

The architecture development underwent three experimental phases to optimize the prediction accuracy. Initially, the baseline $1$NN$2$out architecture (junction temperature and hotspot position are output simultaneously after the fourth hidden layer) demonstrated compromised accuracy due to conflicting gradient updates between junction temperature and hotspot position estimation tasks. This prompted the $2$NN$2$out investigation, where single networks achieved superior precision. Subsequent $1$NN$2$out adopted the architectural refinement which put the junction temperature prediction right after the first hidden layer. Training data (inputs and labels) for both models consist of material and structural parameters, including features such as the thickness of each chip layer, the number of layers, thermal conductivity, power and HTC, as well as simulation results.

To quantify the prediction accuracy of our models, the mean squared error (MSE) loss function is employed. Specifically, the loss function for predicting junction temperature is defined as:
\begin{equation}\label{eqn-8} 
	\text{MSE$_T$} = \frac{1}{N} \sum_{i=1}^{N} (T_i - \hat{T}_i)^2
\end{equation}
where \( T_i \) represents the actual temperature value for the \(i\)-th sample, and \( \hat{T}_i \) is the predicted temperature value from the model for the same sample. Similarly, for the junction position prediction, the loss function is defined as:
\begin{equation}\label{eqn-9} 
	\text{MSE$_P$} = \frac{1}{N} \sum_{i=1}^{N} (P_j - \hat{P}_j)^2
\end{equation}
where: \( P_j \) represents the actual position value for the \(j\)-th sample, and \( \hat{P}_j \) is the predicted position value by the model, \(N\) is the total number of samples used in the training datasets. 

\begin{equation}\label{eqn-10} 
	\text{MSE$_{1NN2out}$} = \text{MSE$_T$} + 10*\text{MSE$_P$} 
\end{equation}
MSE$_{1NN2out}$ is calculated by adding MSE$_T$ and MSE$_P$, specifically by assigning a coefficient of 10 to MSE$_P$.  The advantage of this operation is that it aids in achieving a faster convergence of the position. By minimizing the MSE$_{1NN2out}$, the neural network is trained to refine its parameters so that its predictions for both junction temperature and hotspot position are as close as possible to the true values. 

The GELU activation function \cite{hendrycks2016gaussian}, based on the Gaussian distribution, has been chosen due to its smooth and non-monotonic nature, which helps avoid the vanishing gradient problem during training. Its mathematical expression is given by:
\begin{equation}\label{eqn-7} 
\text{GELU}(x) = x \cdot \frac{1}{2} \left[1 + \frac{2}{\sqrt{\pi}} \int_0^{\frac{x}{\sqrt{2}}} e^{-t^2} \, dt \right]
\end{equation}

\subsection{Training and parameter adjustment}

\begin{figure}
	\centering
	\includegraphics[width=0.45\textwidth]{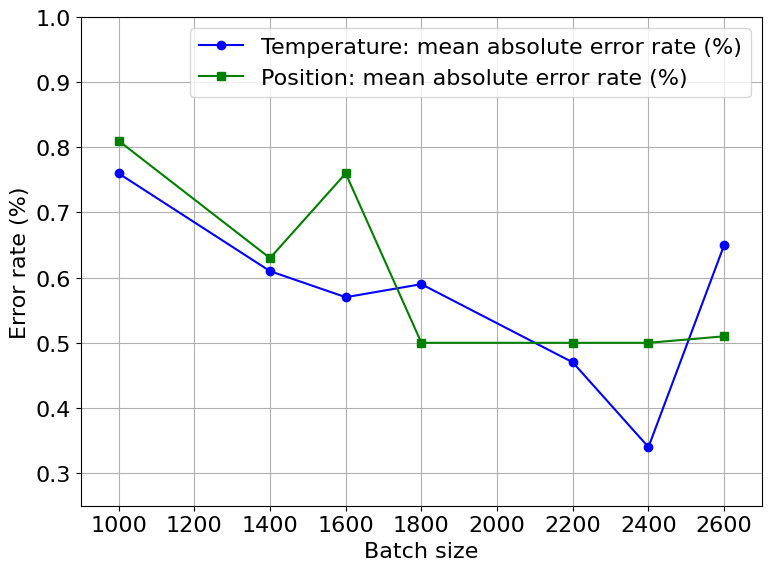}
	\caption{Mean error rate across different batch sizes.}
	\label{pic:G}
\end{figure}

\begin{table}[!h]
	\centering
	\caption{Performance metrics for 2NN2out and 1NN2out network structures.}
	\label{tab:example}
	\setlength{\tabcolsep}{0.15mm} 
	\begin{tabularx}{\linewidth}{llc*{4}{>{\centering\arraybackslash}X}} 
		\hline
		\textbf{net} & \textbf{parameters} &  & \textbf{max-E} & \textbf{MAE} & \textbf{MAR} \\  \hline
		
		\multirow{2}{*}{1NN2out} 	& \multirow{2}{*}{239,106}	& T &  1.57$\tccentigrade $ 			& 0.37$\tccentigrade $  			& 0.34\%  \\
		&														& P &  66.99 \textmu m 	& 0.96 \textmu m 	& 0.45\%  \\
		
		\multirow{2}{*}{2NN2out} 	&	238,658	& T  & 1.23$\tccentigrade $ &0.36$\tccentigrade $  & 0.41\%  \\
		& 243,418	& P & 46.4 \textmu m& 1.29 \textmu m&0.63\%   \\
		\hline
	\end{tabularx}
\end{table}

The adam optimizer is used to optimize both models. For $1$NN$2$out model, the learning rate is initially set to $0.001$ and follows a StepLR, where the learning rate is reduced by a factor of $0.5$ every $5000$ epochs. The batch size is set to $2400$, and the total number of epochs is $200,000$. Training is carried out on NVIDIA GeForce RTX $2060$. The choice of batch size significantly impacts both training efficiency and model performance. As shown in fig. \ref{pic:G}, the selected batch size strikes a balance between training efficiency and model accuracy, ensuring robust performance while maintaining reasonable training times.

\begin{table}[htbp]
	\centering
	\caption{Simulation and inference timing metrics.}
	\label{tab:timing metrics}
	\setlength{\tabcolsep}{8mm} 
	\begin{tabular}{llllll}
		\hline
		\textbf{Metric} 			& {\textbf{Time}} \\
		\hline
		Total simulation (13494)  	&    160 h\\
		\quad \quad 1HBM& 32.20s*504\\
		\quad \quad 2HBM & 37.00s*5832\\
		\quad \quad 4HBM & 44.20s*3391\\
		\quad \quad 8HBM (LHS)& 53.80s*3767\\
		Total inference  &    1.8 s\\
		\quad \quad Per inference case 			&    0.0009s\\
		\hline
	\end{tabular}
\end{table}

For $2$NN$2$out model, the learning rate is set to a constant value of $0.001$. The batch size is set to $1000$, and the total number of epochs is $3000$ for junction temperature prediction and $100,000$ for hotspot position prediction. Tab. \ref{tab:example} compares the performance metrics of the 1NN2out and 2NN2out network structure. In addition to the number of trainable parameters, three error metrics are used for evaluation: maximum error (max-E), mean absolute error (MAE) and mean absolute error rate (MAR). The results demonstrate that the $1$NN$2$out architecture not only reduces model complexity by nearly half, but also achieves better performance in most metrics. However, the $2$NN$2$out model exhibits limitations in training efficiency compared to the $1$NN$2$out model, primarily due to its requirement of repeatedly processing the same input data separately for predictions for junction temperature and hotspot position. This redundant input handling increases computational overhead and training time, making the model less efficient. 

Tab. \ref{tab:timing metrics} provides a detailed overview of the timing metrics for simulation, NN training, and inference. The FEM simulation to generate the training data is the most time-consuming phase. In contrast, the NN training phase demonstrates significant efficiency, which is completed in a relatively short time frame despite the large datasets and the large number of training epochs. The inference phase further underscores the model's computational efficiency, and the per-case inference time being almost negligible. This efficiency makes the model highly suitable for fast and real-time predictions in the early-design stage.

\section{Results and analysis}

This section is divided into two parts: the first subsection involves training the model on the full datasets with ($13,494$ samples), while the second part explores training on a partial datasets with $2,000$ data randomly sampled from the full datasets. The results demonstrate that the model's performance is nearly identical when trained on both the full datasets and the part datasets, indicating that appropriately reducing the data volume has a negligible impact on model accuracy. The results demonstrate that the model trained on the partial datasets performs slightly worse than the one trained on the full datasets, but it is still sufficient to accurately predict the junction temperature and hotspot position.

\subsection{Full datasets}

\begin{figure}
	\centering
	\includegraphics[width=0.5\textwidth]{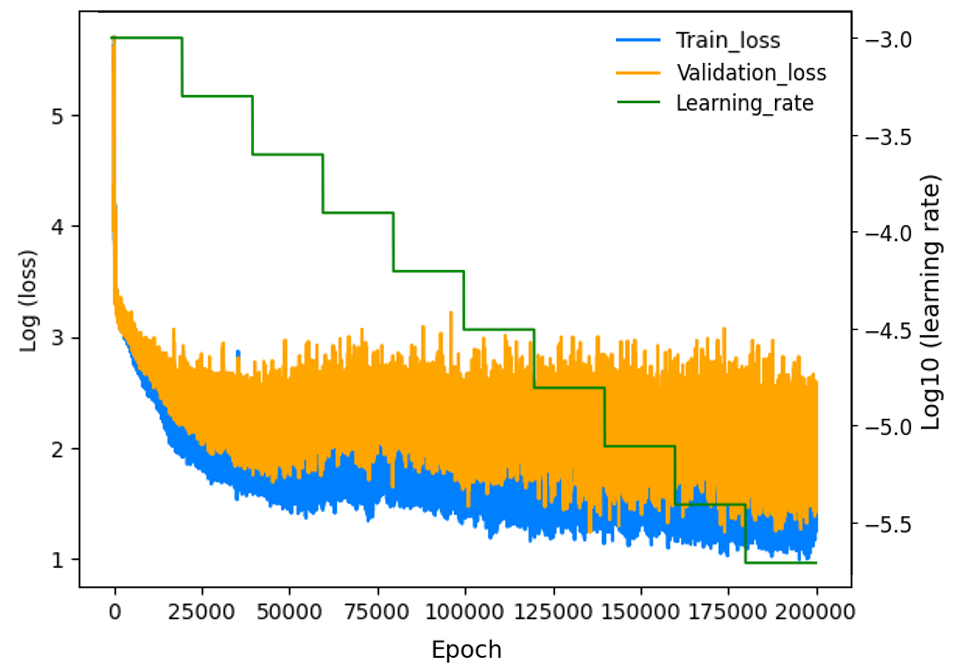}
	\caption{Neural network training loss, validation loss and learning rate curve.}\label{pic:5}
\end{figure}

\begin{figure}
	\centering
	\includegraphics[width=0.5\textwidth]{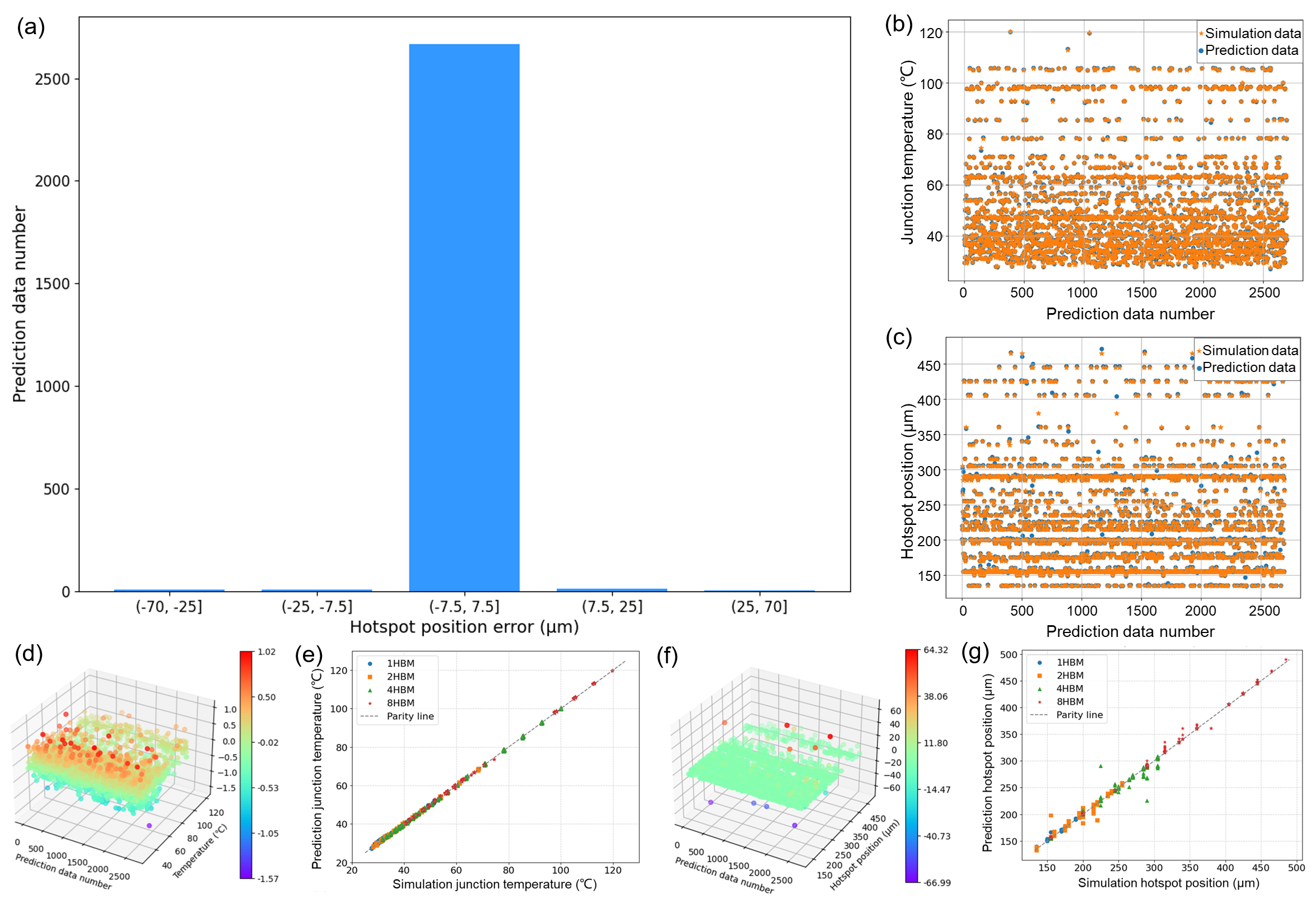}
	\caption{Comparison of predicted and simulated results for full datasets: (a) hotspot position error magnitude distribution, (b) junction temperature scatter plot, (c) hotspot position scatter plot, (d) junction temperature error distribution and (e) predicted-simulated correlation, (f) hotspot position error distribution and  (g) predicted-simulated correlation.}
	\label{pic:X}
\end{figure}

Fig. \ref{pic:5} shows progression of neural network training and testing losses, indicating convergence. Fig. \ref{pic:X} (b) and (c) shows the test results on the test data of the full datasets, where the x-axis represents the prediction data number. The predicted data (blue circles) exhibit strong agreement with simulated data (orange stars) in most cases. However, some discrepancies are observed in the hotspot position predictions. Fig. \ref{pic:X} (d) and (f) show the error of the predicted junction temperature and the hotspot position. The model has high accuracy in predicting junction temperature, with a maximum error of $2.03\tccentigrade $, a mean absolute error of $0.34\tccentigrade $. For hotspot position predictions, the maximum error is $64.32$ \textmu m, with a mean absolute error of $1.01$ \textmu m. These results demonstrate that the model performs well overall, with only a small number of data showing anomalies in hotspot position predictions, while still maintaining acceptable accuracy. As shown in fig. \ref{pic:X} (e) and (g), the junction temperature points overlap almost perfectly with the reference line (y = x), confirming the model accuracy. In contrast, the hotspot position predictions show slightly more variability, with some points scattered on both sides of the reference line. 

Further analysis of these position prediction errors is shown in fig. \ref{pic:X} (a), where the x-axis represents the error range. The interval of $7.5$ \textmu m is chosen because the minimum distance between the adjacent layers of the top surface is $15$ \textmu m, while the interval of $25$ \textmu m corresponds to the minimum distance between other chips. The majority of data ($98.41$\%) falls within the $7.5$ \textmu m range, with only a small portion ($1.22$\%) in the $25$ \textmu m range. Notably, an extremely small fraction ($0.37$\%) of the predictions results in layer misplacement, indicating a minimal impact on overall performance. 

\begin{figure}
	\centering
	\includegraphics[width=0.40\textwidth]{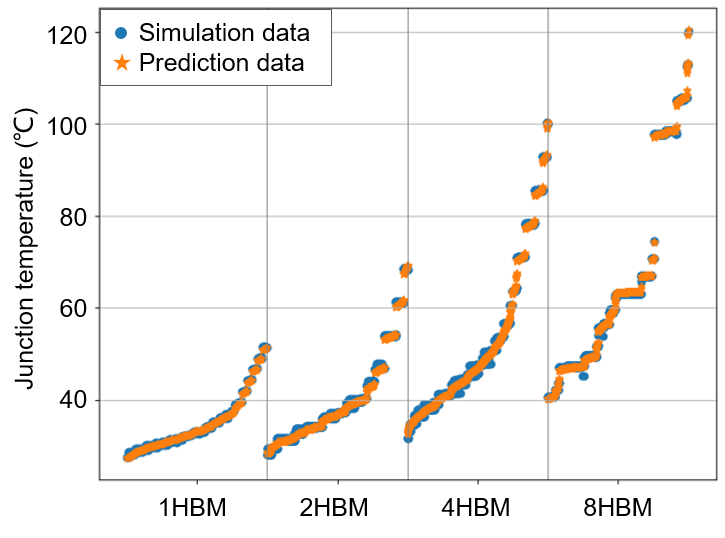}
	\caption{predicted versus simulated junction temperature for 1HBM, 2HBM, 4HBM, and 8HBM, with regions scaled equally to accommodate varying data point densities.}
	\label{pic:B}
\end{figure}

\begin{table}[htbp]
	\centering
	\caption{The performance of temperature and position predictions for $4$-type chiplet ($1$HBM, $2$HBM, $4$HBM, and $8$HBM).}
	\label{tab:summary}
	\setlength{\tabcolsep}{3mm} 
	\begin{tabular}{lllll}
		\toprule
		\textbf{Metric} & \textbf{1HBM} & \textbf{2HBM} & \textbf{4HBM} & \textbf{8HBM}\\
		\midrule
		T$_{MAE}$ ($\tccentigrade $) & 0.71 & 0.30 & 0.58 & 0.29 \\
		T$_{MAR}$ (\%) & 0.85 & 1.01 & 1.05 &0.45 \\
		P$_{MAE}$ (\textmu m) & 0.62 & 1.05 & 2.11 &1.34 \\
		P$_{MAR}$  (\%) & 0.38 & 0.54 & 0.90 &0.49 \\
		\bottomrule
	\end{tabular}
\end{table}

Fig. \ref{pic:B} shows the temperature distribution along the vertical direction for $1$ HBM, $2$ HBM, $4$ HBM, and $8$ HBM, where each vertical sub-region corresponds to one HBM chip. One can see that the temperature range expands as the number of layers increases due to the accumulation of heat from additional heat sources and the increased thermal resistance. The temperature typically is the lowest at the bottom surface (left line of each sub-region) and gradually increases to the top surface (right line of the each sub-region). Tab. \ref{tab:summary} provides a detailed comparison of the performance for temperature and position predictions across the four chip types ($1$HBM, $2$HBM, $4$HBM, and $8$HBM). Overall, the model achieves high accuracy in temperature prediction, with very low average temperature errors across all layers. For position prediction, most errors are relatively small.

\begin{figure}
	\centering
	\includegraphics[width=0.43\textwidth]{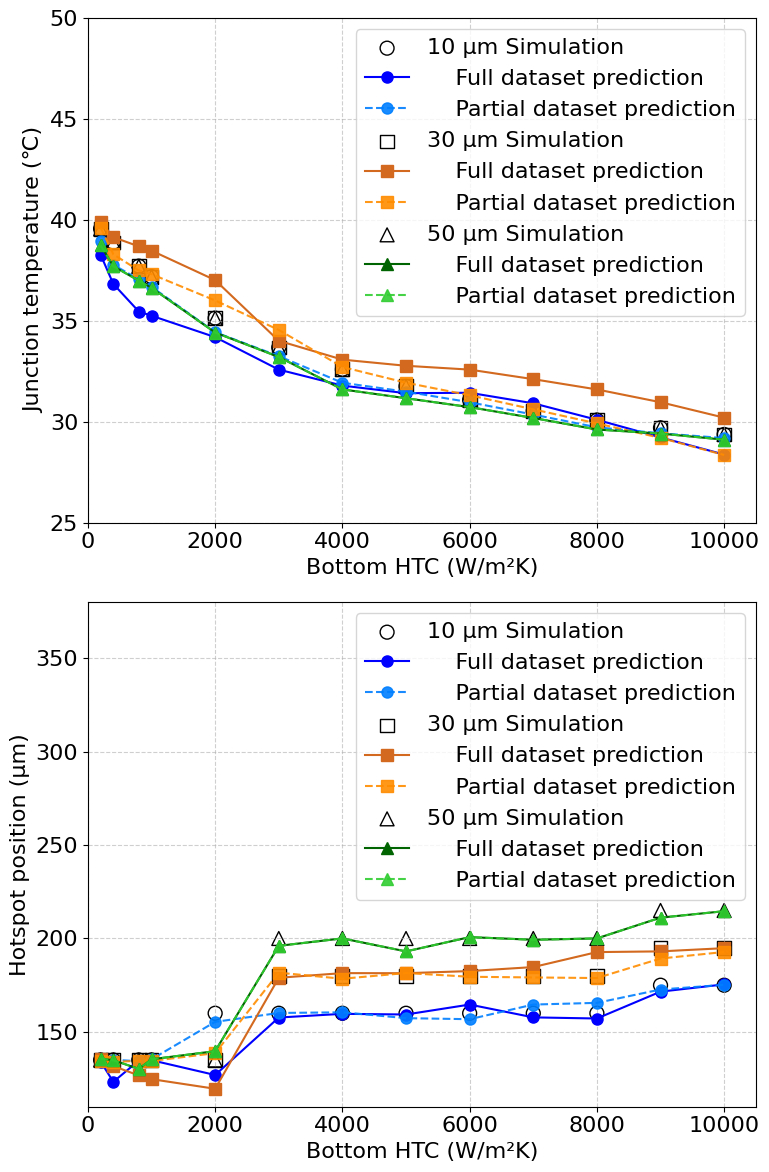}
	\caption{Junction temperature (top) and hotspot position (bottom) with HTC for chip thicknesses.}
	\label{pic:R}
\end{figure}

\begin{figure}
	\centering
	\includegraphics[width=0.42\textwidth]{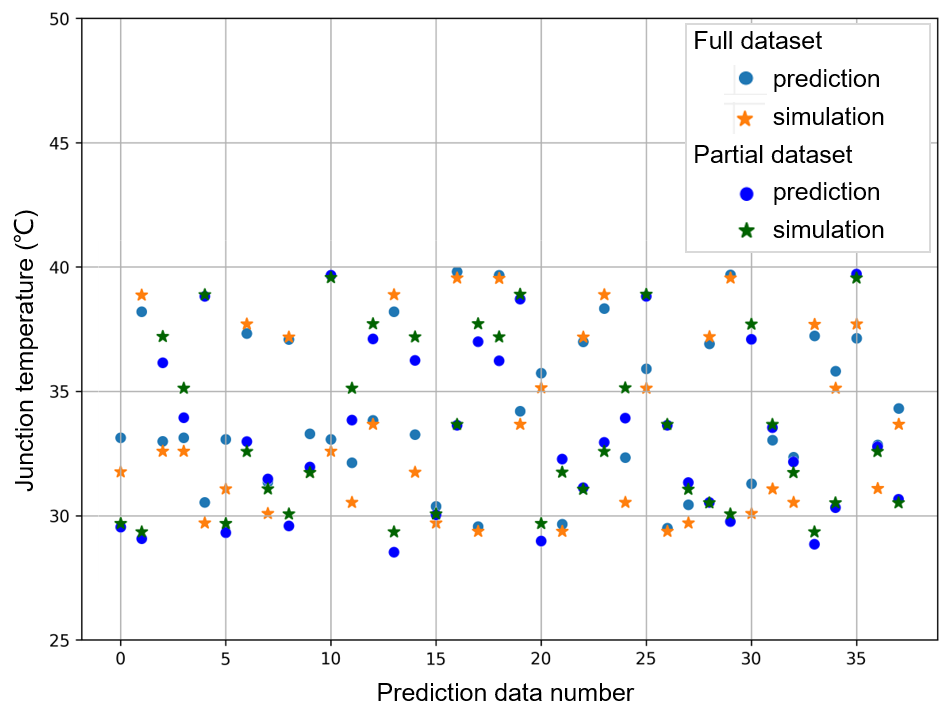}
	\includegraphics[width=0.45\textwidth]{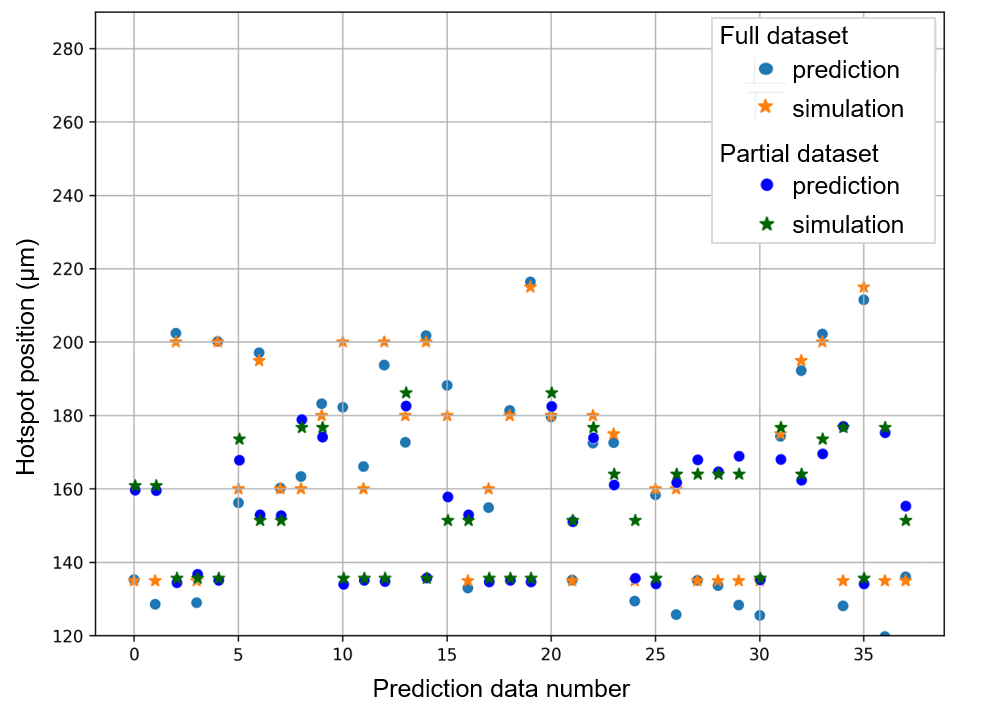} 
	\caption{Scatter plot of junction temperature (top) and hotspot position (bottom) with HTC for chip thicknesses.}\label{pic:I}
\end{figure}

As shown in fig. \ref{pic:R} (top), the junction temperature decreases with increasing HTC from the full datasets, indicating improved heat dissipation at higher HTC values. Thicker chips exhibit higher junction temperatures because of increased thermal resistance, while thinner chips benefit more significantly from enhanced HTC. Fig. \ref{pic:R} (bottom) shows the variation in hotspot position with HTC, where the position shifts slightly as HTC changes, reflecting the impact of thermal gradients on hotspot position. Fig. \ref{pic:I} presents a comparison of prediction results based on the non-sampling points in the full-dataset, these additional HTC values were used to evaluate the model's generalization capability. The figure depicts the junction temperature (top) and hotspot position (bottom), demonstrating the model's ability to accurately predict the thermal behavior for unseen HTC values. By comparing the data before and after the prediction, the model's performance across different chip thicknesses demonstrates its generalization capability. The results indicate that the model can predict the trend of junction temperature and position at non-trained sample points with reasonable accuracy.

\begin{figure}
	\centering
	\includegraphics[width=0.40\textwidth]{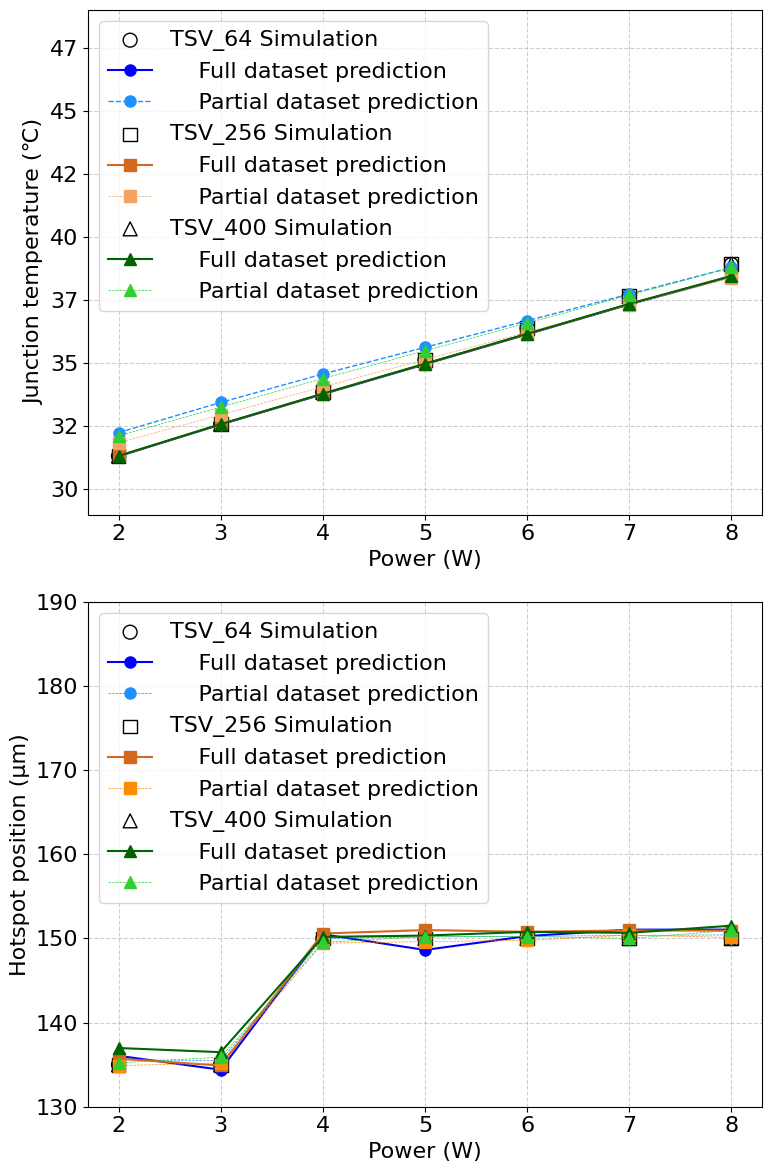}
	\caption{Junction temperature (top) and hotspot position (bottom) with power for chip \(N_{tsv}\).}
	\label{pic:T}
\end{figure}

Fig. \ref{pic:T} shows the variation of junction temperature with power for chips with different TSV numbers (\(N_{tsv}\)) from the full datasets. As power increases, the junction temperature increases, reflecting the linear relationship between power dissipation and thermal load. Chips with higher $N_{tsv}$ values exhibit lower junction temperatures due to improved heat conduction through the additional TSV, which improves thermal dissipation. In contrast, chips with fewer TSV experience higher temperatures and greater variability in hotspot position under the same power conditions, as localized heat accumulation becomes more pronounced. These results demonstrate that increasing $N_{tsv}$ not only reduces the junction temperature but also stabilizes the location of hot spots, highlighting the dual benefits of a higher TSV density for thermal management.

\subsection{Partial datasets}

\begin{figure}
	\centering
	\includegraphics[width=0.40\textwidth]{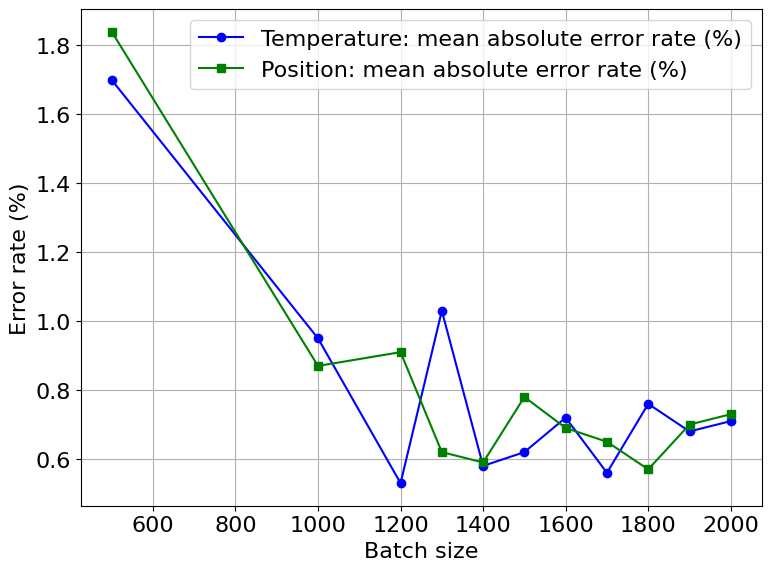}
	\caption{Partial datasets: mean error rate across different batch sizes.}
	\label{pic:J}
\end{figure}

\begin{figure}
	\centering
	\includegraphics[width=0.5\textwidth]{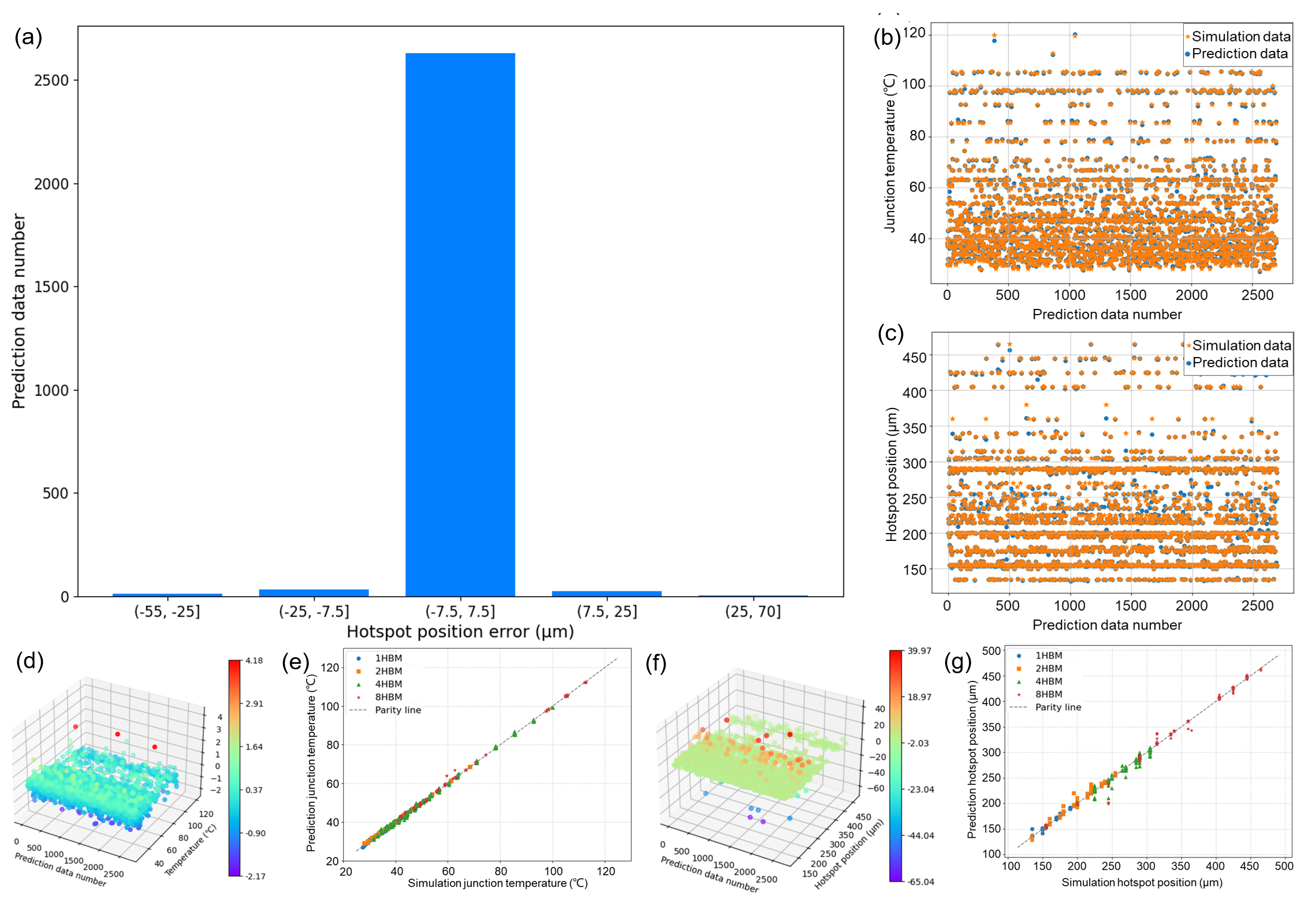}
	\caption{Comparison of predicted and simulated results for partial datasets: (a) hotspot position error magnitude distribution, (b) junction temperature scatter plot, (c) hotspot position scatter plot, (d) junction temperature error distribution and (e) predicted-simulated correlation, (f) hotspot position error distribution and  (g) predicted-simulated correlation.}
	\label{pic:6}
\end{figure}

As shown in fig. \ref{pic:J}, the selected batch size of $1400$ strikes a balance between training efficiency and model accuracy. In fig. \ref{pic:6} (b) and (c), when the datasets is reduced to $2,000$ samples, the model achieves a MAR of 0.58\% for temperature and 0.59\% for position. In contrast, with the full datasets of $13,494$ samples, the mean temperature MAR is 0.34\% and position MAR is 0.45\%. This indicates that the model can still achieve relatively high accuracy with significantly less data, demonstrating its robustness and efficiency. However, when the datasets size falls below $2,000$, the training performance degrades, as the critical patterns or edge cases may no longer be adequately represented. The error analysis in fig. \ref{pic:6} (d) and (f) visualizes the spatial distribution of the prediction errors in the HBM structure for the partial datasets. The results demonstrate that the model achieves satisfactory accuracy with partial datasets, as the error distribution remains relatively uniform and within acceptable bounds. This further supports the model's robustness and efficiency in handling reduced datasets while maintaining reliable performance. Fig. \ref{pic:6} (e) and (g) compares the predictions of neural networks with the finite element simulation results for both the junction temperature and the position of the hotspot in HBM architectures. A good alignment is observed for the junction temperature, with data points tightly clustered along the parity line. For the hotspot position prediction, while a significant majority of neural network predictions correlate well with simulation, some deviations can be observed.  

Further analysis of these mismatches is shown inset in fig. \ref{pic:6} (a), where the x-axis represents the error range. The majority of data ($97.44$\%) falls within the $7.5$ \textmu m range, with only a small portion ($2.15$\%) in the $25$ \textmu m range. In particular, an extremely small fraction ($0.41$\%) of the predictions results in layer misplacement, indicating a minimal impact on overall performance. The overall results show a marginal decline in performance relative to the model trained with full datasets. Nonetheless, they retain a high degree of reliability, indicating that the slight reduction in performance does not significantly affect the general validity of the findings.

As shown in fig. \ref{pic:R}, the junction temperature decreases with increasing HTC from the partial datasets, with thicker chips exhibiting higher temperatures due to increased thermal resistance. The hotspot position changes slightly with HTC, reflecting changes in thermal gradients. In addition to the full-datasets predictions, the performance of the model using partial datasets is also analyzed for comparison. As shown in fig. \ref{pic:I}, predictions based on partial data exhibit little deviations and reduced consistency in capturing the trends of junction temperature and position. Fig. \ref{pic:T} shows that higher \(N_{tsv}\) reduce junction temperature and stabilize hotspot position with increasing power, while fewer TSV lead to higher temperatures and greater position variability. Similarly, reducing the datasets to $2,000$ samples yields a nearly identical performance to the full datasets ($13,494$ samples), indicating that the model can achieve high precision with significantly less data.

This approach not only reduces computational costs and training time, but also provides a practical solution for scenarios where data collection is resource intensive. Overall, the results highlight the model's ability to generalize well even with limited data, making it suitable for real-time applications and early design-phase optimizations. However, in cases where the datasets is significantly larger or more complex, reducing the data volume may introduce greater bias, as critical patterns or edge cases might be underrepresented.

\section{Conclusion}
This study develops a data-driven neural network framework to predict the thermal performance of 3D HBM structures under various thermal conditions. The model can predict junction temperature and hotspot position as functions of key parameters, including chip layer count, layer thickness, material thermal conductivity, power density distribution, and heat exchange coefficients. By implementing neural network inference acceleration techniques, the framework achieves $98.41$ \% prediction accuracy relative to the FEM benchmarks while reducing computational costs. The model demonstrates strong generalization and enables rapid predictions during the early design stage, offering a practical solution for thermal optimization.

\printcredits

\bibliographystyle{unsrt}

\bibliography{cas-refs}

\bio{}
Chengxin Zhang is a Ph.D. candidate jointly affiliated with the School of Microelectronics at Southern University of Science and Technology (SUSTech) and  Peng Cheng Laboratory.  Her research focuses on AI for Science, particularly in use machine learning techniques to accelerate electro-thermal simulations of packaging-related electronic devices.
\endbio

\bio{} 
Dr. Yujie Liu is an Associate Researcher at Peng Cheng Laboratory and PhD supervisor jointly affiliated with Southern University of Science and Technology (SUSTech). She received his Ph.D. in Mathematics from the University of Aix-Marseille in 2014. Her research focuses on numerical and artificial intelligence (AI) methods, high-performance computing and software development for multi-scale multi-physics phenomenon in engineering, electronic design automation (EDA), material, and biomedicine, etc.
\endbio

\bio{} 
Dr. Quan Chen is an Assistant Professor (Associate Researcher) and PhD supervisor at the School of Microelectronics, Southern University of Science and Technology (SUSTech). He received his Ph.D. in Electrical and Electronic Engineering from the University of Hong Kong in 2010 and previously held postdoctoral and research assistant professor positions at the University of California, San Diego (UCSD) and the University of Hong Kong before joining SUSTech in 2019. His research focuses on advanced algorithms and tools in electronic design automation (EDA), particularly in large-scale analog and RF circuit simulation, multi-physics analysis, and nano-device simulation, with significant applications in post-Moore era technologies, including AI-assisted EDA and quantum device modeling.
\endbio

\end{document}